\documentclass[]{elsarticle}
\usepackage{amsmath}
\usepackage{subfigure}
\usepackage{multirow}
\usepackage{url}
\usepackage{color}
\usepackage{graphicx}
\usepackage{longtable}
\usepackage{natbib}

\begin{document}
\begin{frontmatter}
\title{A Benchmark Dataset with Larger Context for Non-Factoid Question-Answering over Islamic Text}

\author{Faiza Qamar}
\address{School of Electrical Engineering and Computer Science (SEECS)}
\address{National University of Sciences and Technology (NUST),\\
        Islamabad, Pakistan.}
\author{Seemab Latif\daggerfootnote{Corresponding author: Seemab Latif (Email: seemab.latif@seecs.edu.pk; ORCID: 0000-0002-5801-1568)}}

\address{School of Electrical Engineering and Computer Science (SEECS)}
\address{National University of Sciences and Technology (NUST),\\
        Islamabad, Pakistan.}

\author{Rabia Latif}
\address{College of Computer and Information Sciences (CCIS)}
\address{Prince Sultan University, Riyadh, 12435, Saudi Arabia}

\begin{abstract}
Accessing and comprehending religious texts, particularly the Quran (the sacred scripture of Islam) and Ahadith (the corpus of the sayings or traditions of the Prophet Muhammad), in today's digital era necessitates efficient and accurate Question-Answering (QA) systems. Yet, the scarcity of QA systems tailored specifically to the detailed nature of inquiries about the Quranic Tafsir (explanation, interpretation, context of Quran for clarity) and Ahadith poses significant challenges. To address this gap, we introduce a comprehensive dataset meticulously crafted for QA purposes within the domain of Quranic Tafsir and Ahadith. This dataset comprises a robust collection of over 73,000 question-answer pairs, standing as the largest reported dataset in this specialized domain. Importantly, both questions and answers within the dataset are meticulously enriched with contextual information, serving as invaluable resources for training and evaluating tailored QA systems. However, while this paper highlights the dataset's contributions and establishes a benchmark for evaluating QA performance in the Quran and Ahadith domains, our subsequent human evaluation uncovered critical insights regarding the limitations of existing automatic evaluation techniques. The discrepancy between automatic evaluation metrics, such as ROUGE scores, and human assessments became apparent. The human evaluation indicated significant disparities: the model's verdict consistency with expert scholars ranged between 11\% to 20\%, while its contextual understanding spanned a broader spectrum of 50\% to 90\%. These findings underscore the necessity for evaluation techniques that capture the nuances and complexities inherent in understanding religious texts, surpassing the limitations of traditional automatic metrics.
\end{abstract}

\begin{keyword}
Question-Answering System; Non-Factoid Question-Answers; Question-Answering Dataset; Automatic Evaluation Techniques; Human Evaluation
\end{keyword}

\end{frontmatter}

\section{Introduction}
In the age of rapidly advancing technology and the increasing reliance on digital resources, there arises a pressing need for efficient and accurate methods to access and comprehend religious texts \citep{MARAOUI202169}. Specifically, the Quran and Ahadith hold immense significance for millions of individuals seeking guidance and understanding in their religious practices \citep{koenig2014health}. The Quran is the central religious text in Islam, revealed upon the last Prophet Muhammad (P.B.U.H). Tafsir, the interpretation of the Quran, helps us understand its meaning and context. On the other hand, Ahadith are the sayings and actions of Prophet Muhammad (P.B.U.H), providing guidance for Muslims \citep{Hamed2016AQA}. Being the fundamental sources of Islamic legislation, having a reliable and comprehensive QA system designed specifically for the Quran and Ahadith can greatly aid in exploring and understanding these important texts. However, exploring these vast repositories of knowledge can be a time-consuming and challenging task \citep{sadek2016discourse}.

By leveraging advancements in natural language processing and machine learning, Question-Answering (QA) systems offer the promise of swiftly retrieving relevant passages and generating answers to questions in natural language \citep{paramasivam2022survey}. In Question-Answering Systems (QAS), Long-form question-answering (LFQA) is an intriguing challenge that involves retrieving documents relevant to a given question and using them to generate a \textit{paragraph-length} answer \citep{dash2023open}. While there has been remarkable recent progress in factoid open-domain question-answering, where a short phrase or entity is enough to answer a question \citep{hao2022motif}, LFQA is still understudied and challenging for Large Language Models (LLMs) \citep{bhat2023investigating}. LFQA is an important task, especially because it provides a test-bed to measure the factuality of generative text models. To make progress in long-form question-answering, researchers need a large, diverse dataset of complex how- and why-type questions with paragraph-length answers.

While QA systems have been developed for various fields, their application to the Quran, Tafsir, and Ahadith is very significant, but scarce \citep{bashir2023arabic}. Muslims worldwide rely on Muslim scholars for guidance in their daily life queries. Several studies have concentrated on a wide range of topics on Islamic text from retrieval to classification. However, where QA is of concern only factoid type has been addressed \citep{hao2022motif}.
Some of these studies are in Arabic \citep[]{MARAOUI202169,malhas2022qur, abdelnasser2014bayan, shmeisani2014semantically, hakkoum2016semantic}, English \citep{Hamed2016AQA}, and Indonesian \citep{putra2016semantic}. Whereas question answers available online on \url{https://islamqa.org}\footnote{It has indexed the answers to over 90K questions of the Muslims across the globe} strongly points out that the users require not only factoid but detailed answers to their questions with references from the Quran and Ahadith. 

A QA system targeting Quran and Ahadith that could answer users' questions with extensive details has its own set of challenges. These challenges include but are not limited to; lack of dataset, appropriate question classification system, accurate facts extraction from different sources while keeping in mind the context provided by the user, to infer the answer \citep{utomo2020question}, and the absence of suitable evaluation techniques that can adequately address the sensitivity of this domain because precision is of utmost importance here. 

This paper has two significant contributions in the following areas:

\begin{enumerate}
    \item The research presents a comprehensive and extensive dataset specifically designed to address the question-answering problem in the domain of Quran, Tafsir, and Ahadith. The dataset consists of over 73,000 question-answer pairs, making it, to the best of our knowledge, the largest reported dataset for long-form question-answering in this domain. Importantly, both the questions and answers come with rich contextual information, providing valuable resources for training and evaluating tailored question-answering systems\footnote{Dataset will be released along with the submission of accepted paper}.
    
    \item The paper introduces a benchmark for evaluating question-answering systems targeted towards the intricacies of the Quran, Tafsir, and Ahadith. This benchmark serves as a standardized evaluation framework, allowing researchers to assess the performance of their models and compare them against existing approaches. It enables the advancement of QA systems specifically tailored for religious literature.
    
\end{enumerate}
The following sections will provide a comprehensive analysis of existing literature (Section 2), present the methodology used for data collection (Section 3), discuss the obtained results (Section 4), analyze findings (Section 5), and conclude with implications and suggestions for future research (Section 6).

\section{Related Work}
In the dynamic field of natural language processing, the pursuit of effective long-form question-answering models is underscored by the critical role played by meticulously curated datasets. This literature review embarks on a dual exploration, focusing on datasets that cater to the broader landscape of long-form question-answering, while concurrently digging into specialized datasets tailored for the unique challenges presented by the interpretation and understanding of Quranic verses and Ahadith. By examining the characteristics, methodologies, and outcomes associated with these datasets, in this section, we aim to provide a nuanced perspective on the advancements made in the development of models capable of comprehending and responding to extended queries, with a specific emphasis on the sacred texts of Islam.

It is divided into three parts; review of language models, available datasets and work on religious scriptures.
\subsection{Language Models}
The introduction of LLMs and transformers has significantly shaped research in LFQA. These models have enabled the development of automated systems that can generate detailed, paragraph-length answers to complex questions, addressing real-world issues such as legal literacy, analyzing public opinion on politics, and information retrieval etc. Several recent studies have proposed novel methodologies and frameworks to improve the performance of LFQA models, addressing challenges such as generating faithful answers with less hallucinated content, evaluating long-form output, and incorporating exemplification in question-answering.

Long-form question-answering (LFQA) research has made progress using large pre-trained models, but a primary challenge remains: generating faithful answers with less hallucinated content. To address this, a recent study proposed an end-to-end framework that jointly models answer generation and machine reading, incorporating fine-grained, answer-related salient information to emphasize faithful facts \citep{su2022read}. This approach achieved state-of-the-art results on two LFQA datasets (ELI5 and MS MARCO), outperforming strong baselines on both automatic and human evaluation metrics. A detailed analysis confirmed the effectiveness of this method in generating fluent, relevant, and faithful answers, contributing to the advancement of LFQA research.

Another research has demonstrated the capabilities of large language models (LLMs) in question-answering and long-form text generation, particularly in few-shot closed-book settings. However, evaluating long-form output remains a challenge. A recent study addressed this issue by combining question-answering with long-form answer generation, leveraging multifaceted questions that require information from multiple sources \citep{amplayo2022query}. The authors introduced query refinement prompts to encourage LLMs to explicitly address question ambiguities and generate comprehensive answers. Experiments on ASQA and AQuAMuSe datasets showed that this approach outperformed fully fine-tuned models in closed-book settings and achieved comparable results to retrieve-then-generate open-book models, highlighting a promising direction for evaluating and improving LLMs' long-form answer generation capabilities.

Exemplification, the process of using examples to clarify complex concepts, is a crucial aspect of long-form question-answering (LFQA). Despite its importance, exemplification in QA has received little computational attention. A recent study addressed this gap by conducting a fine-grained annotation of different example types in three corpora, revealing that state-of-the-art LFQA models struggle to generate relevant examples \citep{wang2022modeling}. Furthermore, standard evaluation metrics like ROUGE were found to be insufficient for assessing exemplification quality. The authors proposed a novel approach, treating exemplification as a retrieval problem, which allows for reliable automatic metrics that correlate well with human evaluation. Human evaluation confirmed that the proposed model's retrieved examples were more relevant than those generated by state-of-the-art LFQA models, highlighting the potential of this approach to improve exemplification in LFQA.

The following section highlights the datasets available to support the LFQA and their importance in its progress.
\subsection{Datasets}
 The exploration and analysis of suitable datasets play a pivotal role in advancing the capabilities of models designed to comprehend and respond to complex queries in extended textual contexts. This section of the literature review looks into the landscape of datasets tailored specifically for Long-Form Question-Answering (LFQA), scrutinizing their characteristics, strengths, and limitations. By navigating through the diverse array of datasets available, we aim to gain a comprehensive understanding of the challenges posed by extended-context questions and the advancements achieved through the utilization of various datasets in the development of robust and nuanced question-answering systems.

LFQA term was formally introduced in 2019 by Facebook when they released the dataset Explain Like I'm Five (ELI5) along with a leader board \footnote{\url{https://facebookresearch.github.io/ELI5/}}. ELI5 stands as the most extensive dataset for the Question-Answering task, comprised of posts and comments from the Reddit forum "Explain Like I'm Five" annotated with explanations for diverse concepts \citep{fan2019eli5}.
ELI5 encompasses both abstractive and extractive answers and is the largest reported dataset with 270K QA pairs for Long-form question-answering. The magnitude of the ELI5 dataset assumes particular importance, as it lays the foundation for the development of models adept at handling a broad array of questions and corresponding answers. This capacity is integral for the practical application of LFQA systems, where real-world scenarios demand a nuanced understanding of diverse topics. The substantial size of the dataset contributes to the robustness of LFQA models, enabling them to navigate and respond effectively to the intricate and varied nature of user queries within the broader context of natural language understanding. However, a notable critique of ELI5 lies in potential inaccuracies and incompleteness within explanations. Given that they are contributed by internet volunteers, leaving room for misinformation. The dataset is derived from user-generated content on the Reddit community, which may contain noise, inaccuracies, or subjective interpretations. This can lead to a lower quality of data, which can negatively impact the performance of LFQA models trained on this dataset \citep{zhuang2024toolqa}. Furthermore, the dataset's origins within the Reddit community may introduce biases, deviating from expert or generalized explanations. This bias should be cautiously considered when utilizing the ELI5 dataset for machine learning models. Despite its status as the largest Long Form Question-Answering dataset, its 81\% train/eval overlap impacts model performance, necessitating careful handling during training and evaluation \citep{krishna2021hurdles}. 

Other datasets existed before ELI5 which also addressed the problem of LFQA, i.e. Microsoft Machine Reading Comprehension (MS MARCO) \footnote{\url{https://microsoft.github.io/MSMARCO-Question-Answering/}} and Natural Questions (NQ) \citep{kwiatkowski2019natural}. MS MARCO is a collection of large-scale datasets focused on machine reading comprehension, question-answering, and passage ranking. It is used for various tasks such as question-answering, natural language generation, passage ranking, keyphrase extraction, crawling, and conversational search.
The MS MARCO datasets are derived from real anonymized Bing user queries and real web documents, making them grounded in real-world problems and providing valuable resources for advancing research in these areas \citep[]{bajaj2016ms, arabzadeh2021ms}. 
It emerges as a substantial contributor to the domain of machine question-answering and passage ranking, garnering attention within the literature for its commendable attributes and acknowledged drawbacks. Noteworthy merits encompass the incorporation of approximately 500,000 \footnote{\url{https://www.sbert.net/examples/training/ms_marco/README.html}} authentic search queries sourced from the Bing search engine, providing a reservoir of real-world queries essential for training models in information retrieval. Furthermore, the dataset distinguishes itself by presenting human-generated answers, a facet that bolsters the overall quality of the dataset. Its sheer scale, constituting a large and diverse collection, proves advantageous for the training and evaluation of machine learning models \citep{bajaj2016ms}. 

However, the dataset is not without its demerits, as scrutinized in the literature. Notably, concerns arise regarding the high level of redundancy present in the dataset, potentially influencing the efficacy of model training and evaluation. Additionally, apprehensions about unfair comparisons surface due to the existence of two distinct corpora within MS MARCO, leading to challenges in result reproduction and the monitoring of state-of-the-art outcomes. Further complicating matters, instances of augmented data introducing leaked relevant information contravene the original guidelines of the dataset. The dataset does not include multi-hop reasoning questions, which are important for evaluating models' ability to reason over multiple pieces of information \citep{lassance2023tale}.

The NQ dataset is a large-scale, real-world dataset for QA research. It consists of anonymized, aggregated queries issued to the Google search engine and is designed to drive research in Natural Language Understanding (NLU) and provide a benchmark for QA systems. In contrast with the MS Marco dataset, it contains 100,000 questions with free-form answers. For each question, annotators are presented with 10 passages returned by the search engine. They are asked to generate an answer to the query or state that the answer is not contained within the passages \citep{kwiatkowski2019natural}. 

\citep{louis2023interpretable} addressed the legal literacy gap by proposing an end-to-end methodology for generating long-form answers to statutory law questions. The approach utilized a "retrieve-then-read" pipeline and was supported by the introduction of the Long-form Legal Question-Answering (LLeQA) dataset, comprising 1,868 expert-annotated legal questions in French. While the results showed promising performance on automatic evaluation metrics, a qualitative analysis revealed areas for refinement. The LLeQA dataset has the potential to accelerate research towards resolving real-world issues and act as a benchmark for evaluating NLP models in specialized domains.

Moreover, the development of the WebCPM dataset for Chinese LFQA introduced a unique feature where information retrieval is based on interactive web search, resulting in a pipeline that generates answers comparable to human-written ones \citep{qin2023webcpm}.

These studies collectively demonstrate the significant impact of Datasets and large language models on shaping research in long-form question-answering, addressing various challenges and advancing the capabilities of automated systems in generating detailed, coherent answers to complex questions \citep{das2022query}.

\subsection{Literature on Religious Scriptures}
This section highlights the work done on the Quran and Ahadith. 

Several studies have addressed the challenge of auto-extracting reliable answers from reference texts, such as constitutions or holy books \citep[]{Handojo2011APLIKASIQA, aharon2022deep,utomo2020question}. Among these texts, the Quran and Ahadith hold particular significance as the holy scriptures of Islam, serving as primary legislative sources for millions of Muslims globally.

An Arabic Question-Answering (QA) system specialized in Islamic sciences, including prophetic tradition (Hadith), Hadith narrator encyclopedia \citep{maraoui2022}, and Quran interpretation (Tafsir), was developed to address the complexity of unstructured information in online databases \citep{MARAOUI202169}. The system's knowledge resource is a normalized database in the Text Encoding Initiative (TEI) standard, and it employs a three-phase method: question analysis, information search, and answer processing. A graphic interface allows user interaction. Experimental results on 100 questions in Hadith, narrator, and Tafsir themes showed a 92\% accuracy rate in generating responses, demonstrating the system's effectiveness in providing accurate answers to factoid questions in Islamic research fields. This study contributes to the development of QA systems for specialized domains and languages.

In a similar vein, \citep{ITJ2022} proposed a question-answering system built on a Hadith knowledge graph to address the limitations of existing digital platforms in answering religious questions. The system utilized the Levenshtein distance function to interpret user questions and Neo4J as the graph database to store the Hadith in a graph format. The results showed that (i) a knowledge graph is suitable for representing the Hadith and performing reasoning tasks, and (ii) the proposed approach achieved a top-1 accuracy of 95\%. This study demonstrates the potential of knowledge graph-based question-answering systems for religious texts, enabling users to seek answers to specific questions and promoting a deeper understanding of Islamic knowledge. Notably, their findings indicated an improvement in the system's results through the utilization of the Levenshtein distance approach \citep{yujian2007normalized}.

The "Qur'an QA 2022" shared task was organized to promote state-of-the-art research in Arabic Question-Answering (QA) and Machine Reading Comprehension (MRC) on the Holy Qur'an, a rich source of knowledge for Muslim and non-Muslim inquirers \citep{malhas2022qur}. The task attracted 13 participating teams, with 30 submitted runs, demonstrating the growing interest in QA and MRC research. This overview paper provides insights into the main approaches adopted by participating teams, highlighting trends and perceptions that characterize the submitted systems. The shared task aims to advance research in Arabic QA and MRC, enabling the development of more accurate and efficient question-answering systems for the Holy Qur'an.

The automatic extraction of reliable answers from religious texts, such as the Qurán, poses a significant challenge for the natural language processing community. Despite its importance, previous research on Question-Answering (Q\&A) from the Qurán is limited and lacks a benchmark for meaningful comparison. Recently, a shared task \citep{malhas2022qur} was organized, providing a dataset of 1,093 question-Qurán passage pairs \citep{ahmed2022qqateam}. A participating system achieved the best scores of 0.63 pRR and 0.59 F1 on the development set and 0.56 pRR and 0.51 F1 on the test set, with an Exact Match score of 0.34, highlighting the difficulty of the task and the need for further research. This study contributes to the development of QA systems for religious texts, enabling users to seek answers to specific questions and promoting a deeper understanding of Islamic knowledge.

Despite progress in Question-Answering (QA) systems, Arabic QA systems face challenges, particularly for the Holy Qur'an, due to limited resources and the difference between Classical and Modern Standard Arabic \citep{mostafa2022gof}. To address this, a Deep Learning-based method was proposed for the Qur'an QA 2022 Shared task, fine-tuning models on large datasets before adapting to the target dataset. This approach achieved promising results, with 66.9\% pRR on the development set and 54.59\% pRR on the test set. This study contributes to the development of effective QA systems for the Holy Qur'an, highlighting the potential of Deep Learning techniques in overcoming the limitations of existing resources.

A concept-based searching tool (QSST) for the Holy Quran was developed to facilitate information retrieval for Quran scholars and Arabic researchers \citep{mohamed2022qsst}. The tool consists of four phases: dataset construction by annotating Quran verses based on the ontology of Mushaf Al-Tajweed, word embedding using Continuous Bag of Words (CBOW) architecture, feature vector calculation for input queries and Quranic topics, and retrieving relevant verses by computing cosine similarity. Evaluation metrics (precision, recall, F-score) showed promising results (76.91\%, 72.23\%, 69.28\%), and expert evaluation by Islamic scholars achieved an average precision of 91.95\%. Comparison with existing tools demonstrated the superior performance of QSST, highlighting its potential for efficient concept-based searching in the Quran.

Arabic Question-Answering Systems (QAS) face challenges due to the complexity of the Arabic language, despite its widespread use by 450 million native speakers \citep{alkhurayyif2023comprehensive}. Current QASs are limited to specific domains, and a comprehensive examination is needed to improve development. While previous studies categorized QASs based on various factors, there is a lack of research on development techniques. This systematic literature review aimed to address this gap by analyzing 40 selected papers from a pool of 617 articles. The findings highlight the importance of datasets and deep learning techniques in enhancing QAS performance. Moreover, the reliance on supervised learning methods hinders QAS performance, and the development of unsupervised QAS using advanced machine learning techniques is encouraged. This review provides valuable insights for developing effective Arabic QAS, aligning with the Saudi Arabian government's push for automation and improved services.

Despite the numerous studies conducted in the past decade on Quran and Hadith texts, a significant research gap remains in the availability of a comprehensive dataset that can effectively harness the power of available pre-trained models for non-factoid \citep{cortes2022systematic} question-answering. Furthermore, there is a notable absence of an established evaluation protocol specifically designed for assessing the performance of such sensitive systems. These gaps highlight the need for further research and development in this area to address the challenges of dataset scarcity and the lack of standardized evaluation methods for Quran and Hadith QA systems.

In summary, this literature review navigates the landscape of LFQA in Natural Language Processing, scrutinizing pivotal datasets like ELI5, MS MARCO, and NQ. Evaluating their strengths and limitations, the review showcases their role in advancing machine learning models for intricate QA tasks. It highlights recent strides driven by LLMs and transformers, unveiling methodologies enhancing LFQA model performance. Additionally, the review sheds light on challenges faced in extracting reliable answers from religious scriptures like the Quran and Ahadith, emphasizing the scarcity of comprehensive datasets and evaluation methods in this domain. The identified gaps urge further exploration and the need for robust datasets and standardized evaluation methods in Quran and Hadith QA systems to propel specialized research in this field.

\section{Task Description and Dataset}
This section provides a comprehensive overview of the data sources and processing procedures employed in this study. The dataset was compiled from multiple reputable sources to ensure diversity and authenticity. Firstly, we collected question-answer pairs from Islamqa.org\footnote{Available at \url{https://islamqa.org/}, Last Visited: 14 June 2023.}, a prominent online Question-Answering platform that hosts an extensive collection of over 90,000 question-answer pairs. This platform allows users to pose questions, which are then answered by Muslim scholars based on Islamic Law, providing a valuable resource for understanding Islamic perspectives.

In addition to the question-answer pairs, we obtained the English translation of the Tafsir of the Holy Quran from Al-Tafsir.com\footnote{Tanwir al-Miqbas min Tafsir Ibn Abbas, available at: \url{https://www.altafsir.com/Books/IbnAbbas.pdf}}, a trusted online resource that offers detailed explanations and interpretations of the Quranic text. This Tafsir provides a deeper understanding of the Quran's meaning and context, which is essential for developing a comprehensive Islamic Question-Answering system.
Furthermore, we acquired English translations of over 33,000 Ahadith from the six major Hadith books known as Sahah-e-Sittah\footnote{More information available at: \\ \url{https://en.wikipedia.org/wiki/Kutub_al-Sitta} and \\ \url{https://islamhashtag.com/the-six-sitta-al-sihah- \\ al-sittah-the-authentic-books-on-hadeeth/}}, which are considered the most authentic and reliable sources of Hadith in the Muslim Community. Relying on Sahah-e-Sittah in our study upholds dataset quality and adheres to esteemed sources, ensuring the development of a robust and accurate Islamic Question-Answering system.

The collected data, comprising question-answer pairs, Tafsir, and Ahadith translations, underwent rigorous processing to prepare it for utilization in sequence-to-sequence pre-trained models. These models were then fine-tuned and evaluated for their performance, as detailed in the subsequent sections. The processing steps involved data cleaning, tokenization, and formatting to ensure compatibility with the pre-trained models. The resulting dataset is a comprehensive and diverse collection of Islamic texts, providing a solid foundation for developing an effective Islamic Question-Answering system.

\subsection{Data Preprocessing}
Data preprocessing is a crucial step in preparing the collected data for use in sequence-to-sequence pre-trained models. After collecting the raw data, we cleaned it to eliminate any duplicate entries and missing data. We also removed the Arabic counterparts of answers provided in both English and Arabic languages.

\begin{table}[ht]
\caption{Topics with their Top Keywords}
{\begin{tabular}{p{0.06\linewidth}  p{0.65\linewidth}p{0.1\linewidth}}
\hline
\textbf{Topic} & \textbf{Top Keywords} & \textbf{Label}\\ \hline
1 &
Prayer, Pray, Read, Recite, Imam, Salaah, Mosque, Perform, Answer, Salah
&
Prayer\\  \hline
2
& Fast, Muslim, Allah, People, Person, Month, Good, Feel, Think, Mean
& Fasting\\ \hline
3
& Hajj, Day, Perform, Umrah, Travel, Babi, Go, Name, Makkah, Rule
& Hajj \\ \hline
4
& Hadith, Prophet, Hair, Bless, Preach, Aalayhi, Narrate, S.A.W, Authentic, Follow
& Hadith \\  \hline
5
& Wear, Water, Cloth, Wash, Ghusal, Rule, Hand, Food, Muslim, Allow
& Daily Life \\ \hline
6
& Wife, Husband, Marry, Family, Year, Mother, Divorce, Marriage, Parent, Live
& Marriage \\ \hline
7
& Father, Child, Wealth, Belong, Children, Son, Need, S.A.W, Alaihi, Salam
& Family \\ \hline
8
& Money, Work, Company, Bank, Year, sell, Pay, Give, Month, House
& Finance \\ \hline
\end{tabular}}

\label{tab:2}
\end{table}

To reduce the search space from the complete text of Quranic Tafsir and Ahadith, we employed Latent Dirichlet Allocation (LDA) \citep{yu2001direct, zoya2022} topic modeling. LDA is a probabilistic model that recognizes topics within a document collection by assuming that each document comprises a blend of various topics, where each topic represents a distribution of words. By applying LDA, we aimed to identify the underlying topics in the dataset and group similar documents together, thereby reducing the search space and making the system more efficient \citep{trastour2016prediction}. This approach enables the model to focus on a subset of relevant texts, rather than the entire corpus when generating answers.

We divided the corpus into eight topics, namely Prayer, Fasting, Hajj, Hadith, Daily Life, Marriage, Family, and Finance, using LDA. Table \ref{tab:2} presents the top words from each topic with an assigned label. By categorizing the dataset into these topics, we significantly reduced the search space, allowing the model to quickly identify relevant texts and generate accurate answers.

To ensure the accuracy of the assigned topic labels, we had three language experts evaluate the assigned topic label for each question from the subset of the complete dataset. The evaluators were crowd workers with expertise in Islamic studies and language. They assessed the relevance of the assigned topic labels and provided feedback on the accuracy of the labels. The detailed results of this evaluation are presented in Table \ref{tab:3}, which shows the relative percentage of questions from the total dataset evaluated by the evaluators and the accuracy of the assigned labels. If a label was deemed incorrect by annotators, they assigned the appropriate label from the eight defined categories, and the decision was taken with a majority vote.

\begin{table}[ht!]
\caption{Results of Evaluation of the Extracted Topics}
\centering{
\begin{tabular} {lcccc}
\hline

\textbf{Topic}
& \textbf{Correct/ Evaluated}
& \textbf{Accuracy}
& \textbf{Total Questions }
& \textbf{Percentage of Total } \\

& & & \textbf{from that Topic}
& \textbf{Questions that were} \\ 
& & & & \textbf{Evaluated} \\ \hline

Finance & 365/425 & 85.88\% & 10,203 & 4\% \\ \hline
Prayer & 341/529 & 64.4\% & 14,883 & 3\% \\ \hline
Hajj & 32/88 & 36.3\% & 3,104 & 2\% \\ \hline
Family & 73/385 & 18.9\% & 9,944 & 3\% \\ \hline
Hadith & 124/162 & 76.5\% & 7,722 & 2\% \\ \hline
Marriage & 360/506 & 71.1\% & 14,983 & 3\% \\ \hline
Fasting & 1/4 & 25\% & 3,291 & 0.1\% \\ \hline
Daily Life & 320/364 & 87.9\% & 12,983 & 3\% \\
\hline
Total & 1616/2463 & 64.6\% & 77113 & 20.1\% \\
\hline
\end{tabular}}

\label{tab:3}
\end{table}

By using LDA and evaluating the assigned topic labels, we ensured that the dataset was well-organized, relevant, and ready for use in training sequence-to-sequence pre-trained models to generate accurate and informative answers to user queries. The reduced search space and accurate topic labels enabled the model to efficiently derive answers by focusing on a subset of pertinent text rather than the entire corpus of Ahadith and Tafsir.

When presented with a question, a Muslim scholar typically begins by referring to the Quran and Ahadith to seek the answer. We focused on the Quran and Ahadith texts and drew context from these to feed our seq-to-seq language models. Ahadith and Ayahs that were categorized in the same topic as a question were stored as context for that question-answer pair. However, the resulting context was too large to process, so we picked the top three Ayahs and top three Ahadith based on soft cosine similarity \citep{novotny2018implementation}.

As a result of this process, the dataset was organized into three distinct columns: Question (representing user queries), Answer (responses from Muslim scholars), and Context (comprising excerpts from Quranic Tafsir and Ahadith that had the same topic as the question). The incorporation of context allowed the model to efficiently derive answers by focusing on a subset of pertinent text rather than the entire corpus of Ahadith and Tafsir. An example from the dataset is provided below, demonstrating how the context column provides relevant information from the Quran and Ahadith to support the answer.
By preprocessing the data in this manner, we ensured that the dataset was well-organized, relevant, and ready for use in training sequence-to-sequence pre-trained models to generate accurate and informative answers to user queries.



\begin{figure}[ht]

\centerline{\includegraphics[width=5in]{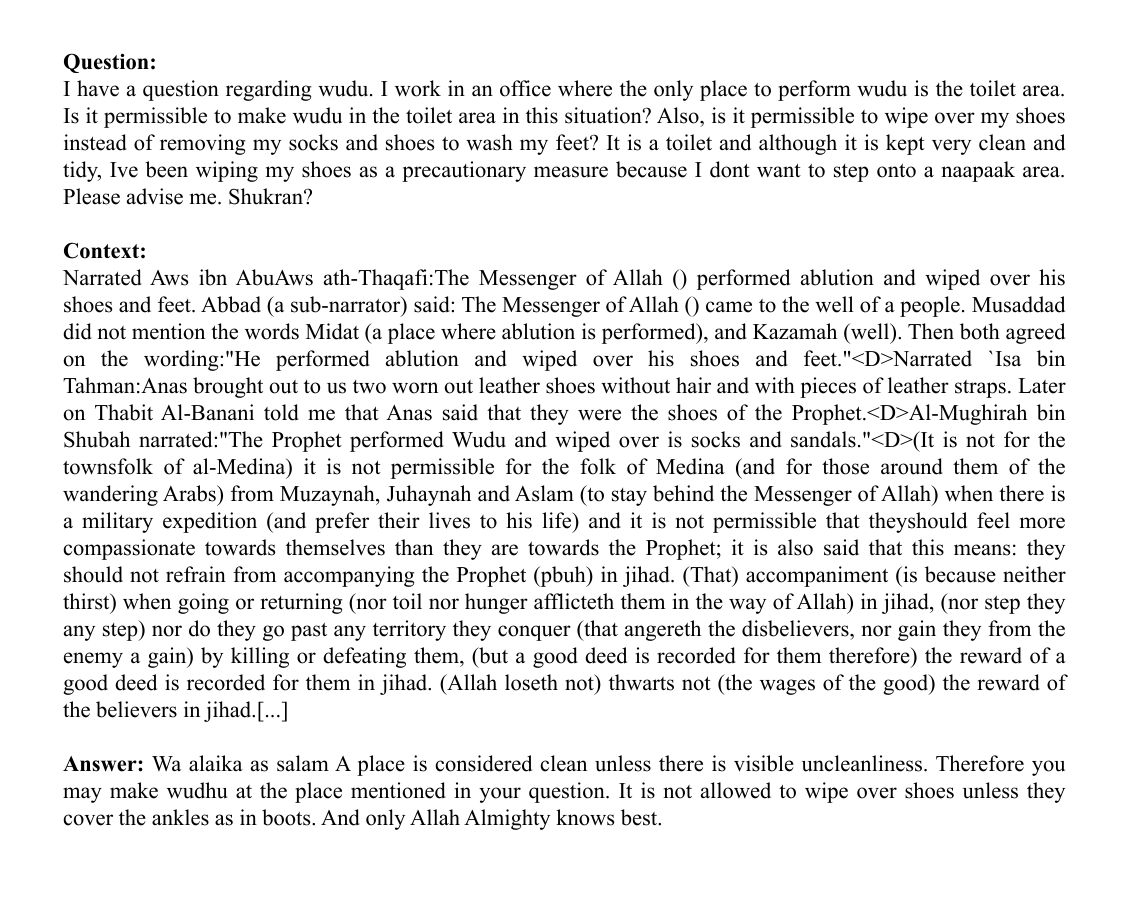}}

\caption{Dataset Example}
\label{fig: t5}
\end{figure}

More examples of this data are available in the appendix \ref{app:dataset} for reference.

\section{Models and Experimental Design}
To establish baseline results for the language models on this dataset, we performed fine-tuning on several transformer-based pre-trained models, including T5, BART, LED, and LongT5. The input format for the models is shown in appendix \ref{app:dataset}, and the hyperparameters and specifications for each model are listed in \ref{tbl:4}.
\subsection{Experimental Setup}
The experiments were conducted using an NVIDIA GeForce GTX 1080 Ti with 12 GB RAM. Due to RAM limitations, the input sequence length and batch size were adjusted accordingly. We observed that not all models performed equally well with the same learning rate. For instance, fine-tuning BART-large with a learning rate of 0.001 (which performed well for T5) resulted in poor performance, significantly worse than any other model used in this study. This contrasts with literature suggesting that BART performs well on language generation and QA tasks, as it was released alongside ELI5 as a state-of-the-art model \citep{lewis2019bart}.
To address these limitations, we used the base versions of the models due to limited RAM space. The hyperparameter settings for each experiment are listed in Table \ref{tbl:4}. We used a batch size of 4 for T5 and LongT5, and 2 for BART and LED, with input sequence lengths adjusted accordingly to fit within the 12 GB RAM constraints.
\subsection{Model Architecture}
The transformer-based pre-trained models used in this study are:
T5: a text-to-text transformer model that generates output text based on input text \citep{roberts2019exploring}.
BART: a denoising autoencoder that uses a transformer encoder and decoder to reconstruct input text \citep{lewis2019bart}
LED: a long-range dependence transformer model designed for long-form text generation \citep{beltagy2020longformer}
LongT5: a variant of T5 designed for long-form text generation \citep{guo2021longt5}
These models were chosen for their ability to handle long-range dependencies and generate coherent text, making them suitable for the non-factoid question-answering task.
\subsection{Hardware Details}
The experiments were conducted on an NVIDIA GeForce GTX 1080 Ti with 12 GB RAM, which provided sufficient computational resources for fine-tuning the pre-trained models. The GPU architecture allowed for parallel processing of the input sequences, enabling efficient training and evaluation of the models.
By using these transformer-based pre-trained models and adjusting the hyperparameters and input sequence lengths to accommodate the RAM limitations, we established baseline results for the language models on this dataset and explored the effectiveness of fine-tuning for non-factoid question-answering tasks.

\begin{table}[ht]

\caption{Experimental Settings for Different Models}
\centering
{\begin{tabular}{p{0.21\linewidth}p{0.13\linewidth}p{0.15\linewidth}p{0.09\linewidth}p{0.07\linewidth}} \hline
\textbf{Model} & \textbf{Input Length} & \textbf{Output Length} & \textbf{learning rate} & \textbf{Batch Size} \\ \hline
t5-base & 512 & 512 & 0.001 & 6 \\\hline
facebook/bart-base-cnn & 1024 & 512 & 2e-5 & 4 \\ \hline
allenai/led-base-16384 & 4096 & 512 & 2e-5 & 3 \\ \hline
google/long-t5-tglobal-base & 4096 & 512 & 1e-3 & 1 \\ \hline
\end{tabular}}

\label{tbl:4}
\end{table}

\subsection{Evaluation Metric}
The evaluation of results was carried out using two complementary metrics: ROUGE \citep{lin2004rouge} and BERTScore \citep{zhang2019bertscore}. These metrics were chosen for their ability to assess the quality and similarity of generated text, and their suitability for evaluating non-factoid question-answering tasks.
\subsubsection{ROUGE}
ROUGE (Recall-Oriented Understudy for Gisting Evaluation) is a widely used metric for evaluating text generation tasks, including machine translation, summarization, and question-answering \citep{lin2004rouge}. It measures text similarity using N-gram overlap, which calculates the number of overlapping words or phrases between the generated text and the reference text. ROUGE is adaptable to various language generation tasks and has been used in numerous studies (Gao et al., 2019; Lewis et al., 2019). We used ROUGE as our evaluation metric to maintain consistency with previous work in the field.
\subsubsection{BERTScore}
BERTScore, on the other hand, is a more recent metric that has been shown to outperform other commonly used metrics such as BLEU and ROUGE on several benchmark datasets \citep{zhang2019bertscore}. It is based on the BERT (Bidirectional Encoder Representations from Transformers) model, a pre-trained neural network that can be fine-tuned for various natural language processing tasks (Devlin et al., 2018). BERTScore computes a similarity score between the generated text and the reference text by comparing their BERT embeddings, which are high-dimensional vectors that represent the meaning of the text. The score is computed at the word, sentence, and document levels, and takes into account both precision and recall  \citep[]{trastour2016prediction,eberts2019span}.

The use of BERTScore is motivated by its ability to capture the semantic meaning of text, rather than just relying on surface-level similarity. This is particularly important for non-factoid question-answering tasks, where the generated text needs to convey the correct meaning and context. By using both ROUGE and BERTScore, we can evaluate the quality of the generated text from different perspectives and gain a more comprehensive understanding of its strengths and weaknesses.
The next section presents and discusses the results in detail, highlighting important future work and potential avenues for improvement.

\section{Results and Discussion}
The fine-tuning of language models on the dataset resulted in improved performance, as evident by the evaluation using ROUGE in Table \ref{tab:5}. It lists the ROUGE difference before and after finetuning the models. Before fine-tuning, the baseline ROUGE scores for all models were relatively low, ranging from 13.5 to 19.25. However, after the fine-tuning, significant enhancements were observed across all models, with ROUGE scores ranging from 24.70 to 27.23. This indicates that the models generated more accurate and relevant text, capturing the underlying concepts present in the dataset. 

\begin{table}[ht]
\caption{Results of Different Models before and after Finetuning}
\label{tab:5}
\begin{tabular}{l|llc|lllc}
\hline
       & \multicolumn{3}{c|}{\textbf{Before Finetuning}}                             & \multicolumn{4}{c}{\textbf{After Finetuning}}                                                                \\ \hline
Model  & \multicolumn{1}{c}{Rouge1} & \multicolumn{1}{c}{Rouge2} & RougeL & \multicolumn{1}{c}{Rouge1} & \multicolumn{1}{c}{Rouge2} & \multicolumn{1}{c}{RougeL} & BERTScore \\ \hline
T5     & \multicolumn{1}{c}{13.87}  & \multicolumn{1}{c}{1.81}   & 9.76   & \multicolumn{1}{c}{26.85}  & \multicolumn{1}{c}{7.33}   & \multicolumn{1}{c}{17.84}  & 78.05      \\ 
BART   & \multicolumn{1}{c}{19.25}  & \multicolumn{1}{c}{2.36}   & 11.92  & \multicolumn{1}{|c}{24.42}  & \multicolumn{1}{c}{5.82}   & \multicolumn{1}{c}{14.52}  & 79.19        \\ 
LED    & \multicolumn{1}{c}{13.5}   & \multicolumn{1}{c}{2.14}   & 7.96   & \multicolumn{1}{c}{27.23}  & \multicolumn{1}{c}{7.55}   & \multicolumn{1}{c}{18.13}  & 78.95      \\ 
LongT5 & \multicolumn{1}{c}{18.35}  & \multicolumn{1}{c}{1.59}   & 10.19  & \multicolumn{1}{c}{24.89}  & \multicolumn{1}{c}{6.64}   & \multicolumn{1}{c}{17.19}  & 80.65     \\ \hline
\end{tabular}
\end{table}


The findings demonstrate the effectiveness of fine-tuning transformer-based language models for the question-answering task using the Quran, Ahadith, and Tafsir datasets. These results provide valuable insights into the potential of using these language models for enhancing QA systems related to Islamic literature. However, we made some interesting observations by conducting a manual analysis of some generated answers which are as follows:

In some answers, the high ROUGE score was not a good indicator of the correct answer, e.g. a question where the user is trying to know if something is permissible or not according to the Islamic Shariah, the ROUGE score could be higher regardless of the generated answer to be correct or not. The answer could be the opposite of the ground truth, factually, but still would have a higher score.

The generated answer was not always a reflection of the context from which the model generated the answer. This suggests that the model also relied on the pre-consumed knowledge on which it was initially trained. However, it can be further verified by conducting a specific study. Furthermore, ROUGE is a recall-oriented matrix. Whereas in this particular domain, achieving the best results requires a balance between both recall and precision. So that it does not miss some important facts (focusing on Recall) and also doesn't infer wrong conclusions (focusing on precision) from those facts. 

\subsection{Human Evaluation}
The human evaluation of the QA system for the Quran and Ahadith was conducted based on two crucial parameters: Verdict Consistency and Contextual Understanding. A team of expert evaluators, comprising Islamic scholars and language experts, assessed the system's performance in providing accurate and consistent verdicts and its ability to comprehend the context of the questions.

\subsubsection{Verdict Consistency}
The assessment focused on determining whether the verdict provided by the scholar and the model coincided. Results showcased a significant discrepancy, with the range of consistency falling notably lower, ranging between 11\% to 20\% (Table \ref{tab:6}). This suggests that the agreement between the model's verdict and the scholar's verdict was substantially inconsistent across evaluations. The highest consistency was achieved by the LED model (22\%), while the LongT5 model showed the lowest consistency (11\%).
\subsubsection{Contextual Understanding}
Another key parameter was the system's ability to comprehend the context of the question and deliver the relevant answer. In this aspect, the results demonstrated a comparatively broader range, spanning between 50\% to 90\% for four models (Table \ref{tab:6}). This indicates that the models' capability to grasp the context varied significantly, with some models showing promising understanding while others fell short. The LED model achieved the highest score (90\%), while the LongT5 model showed the lowest score (53\%).
\begin{table}[ht]
\caption{Results of Human Evaluations}
\label{tab:6}
\centering
\begin{tabular}{lcc}
\hline
\textbf{Model}  & \textbf{Verdict Consistency} & \textbf{Contextual Understanding} \\ \hline
\textbf{T5}     & 0.22                                              & 0.78                                                   \\ \hline
\textbf{BART}   & 0.20                                              & 0.86                                                   \\ \hline
\textbf{LongT5} & 0.11                                              & 0.53                                                   \\ \hline
\textbf{LED}    & 0.22                                              & 0.90                                                   \\ \hline
\end{tabular}
\end{table}

\subsubsection{Discussion}
The evaluation reveals a substantial disparity between the two parameters. While the system exhibited a wider range of context understanding, its consistency in providing verdicts aligned with the scholars remained notably low. This underscores the need for further improvements, particularly in refining the model's ability to produce more consistent and accurate verdicts in alignment with expert scholars while maintaining a consistently high level of contextual understanding.
The human evaluation results highlight the challenges of developing a QA system for non-factoid questions in the domain of Quran and Ahadith. The system's ability to comprehend the context of the question and provide accurate verdicts is crucial for its reliability and trustworthiness. Future work should focus on addressing the inconsistencies in verdict consistency and further enhancing the system's context understanding capabilities.

While ROUGE and BERTScore are widely used metrics for evaluating text generation tasks, they may not directly measure verdict consistency between the model and scholars. This is because they primarily assess the similarity of generated text to reference text, rather than evaluating the accuracy or consistency of the verdicts themselves. However, lower scores on these metrics might indicate divergence in the veracity of answers provided by the model compared to the scholars' answers.
The low ROUGE score, in particular, weakly implies that the model's generated answers significantly differed from the reference scholar's answers, aligning with the observed low consistency between model and scholar verdicts. This suggests that the model's generated answers may not have accurately captured the nuances and complexities of the scholars' answers, leading to inconsistencies in verdict consistency.

On the other hand, ROUGE and BERTScore can indirectly reflect the model's ability to understand the context by measuring the semantic similarity between the generated answers and reference texts. The higher BERTScore metric suggests that the models captured and expressed the context well, consistent with the broader range (50\% to 90\%) of context understanding observed in the human evaluation. This indicates that the models were able to grasp the context of the questions to some extent, but may have struggled to generate answers that accurately reflected the scholars' verdicts.

Overall, while ROUGE and BERTScore provide valuable insights into the model's performance, they should be used in conjunction with human evaluation to get a more comprehensive understanding of the model's strengths and weaknesses. By combining these metrics with human evaluation, we can gain a better understanding of the model's ability to generate accurate and consistent verdicts, as well as its ability to understand the context of the questions.

\section{Conclusion and Future Work}

In conclusion, this paper has introduced a comprehensive dataset tailored for non-factoid question-answering specifically focused on Quranic Tafsir and Ahadith. The dataset, comprising over 73,000 question-answer pairs, is enriched with substantial contextual information, providing a robust foundation for research and development in this domain. The exploration conducted herein demonstrates the potential of harnessing advanced natural language processing and machine learning techniques to construct QA systems within the realm of Islamic literature. However, the limitations of solely relying on automatic evaluation metrics became apparent, highlighting the need for a nuanced evaluation approach that incorporates human assessment and expert feedback.

The dataset's breadth and depth facilitate the application of various techniques, enabling models to navigate the intricate layers of religious texts. Nevertheless, the identified discrepancies between model-generated responses and expert opinions necessitate a deeper understanding and refinement of these systems to ensure their alignment with scholarly interpretations and contextual fidelity. Therefore, our focus lies in refining and augmenting the evaluation strategies, integrating expert assessments, user feedback, and meticulous analyses of contextual accuracy.

However, while initial automatic evaluations using ROUGE scores provided an initial assessment, our subsequent human evaluation yielded critical insights. The discrepancy between automatic evaluation metrics and human assessment became apparent. The human evaluation uncovered significant disparities: the model's verdict consistency with expert scholars ranged between 11\% to 20\%, while its contextual understanding spanned a broader spectrum of 50\% to 90\%. This disparity elucidates the limitations of solely relying on ROUGE scores, underscoring the necessity for a nuanced evaluation approach.

It became evident that automatic evaluation metrics, while useful, fail to encapsulate the complexities and accuracy required in understanding religious texts. Manual analysis highlighted the imperative need for further investigations into the correctness and contextual alignment of the answers generated by these systems. The identified discrepancies between model-generated responses and expert opinions necessitate a deeper understanding and refinement of these systems to ensure their alignment with scholarly interpretations and contextual fidelity.

Future work should address the limitations of this study, including the incorporation of multiple books for Quranic Tafsir to provide a more comprehensive understanding of the context. Additionally, human expert evaluation should be included to compare and determine which automatic evaluation measure is more correlated and aligned with human assessment. This will enable the development of more sophisticated models and techniques tailored to the intricacies of the Quran, Tafsir, and Ahadith.

Furthermore, future research should explore the following avenues:
\begin{itemize}
    \item Investigating the application of advanced NLP techniques, such as transfer learning and multi-task learning, to enhance the performance of QA systems in this domain.
    \item Developing more nuanced evaluation metrics that can accurately capture the complexities and accuracy required in understanding religious texts.
    \item Exploring the use of additional resources, such as Islamic scholarly articles and books, to further enrich the dataset and improve the performance of QA systems.
    \item Investigating the potential applications of QA systems in Islamic literature, such as assisting scholars and researchers in their work, providing educational resources for students, and enhancing the accessibility of Islamic knowledge.
\end{itemize}

By addressing these limitations and exploring these avenues, we can further advance research in non-factoid question-answering and promote the development of more sophisticated models and techniques tailored to the intricacies of the Quran, Tafsir, and Ahadith. This will ultimately contribute to the advancement of QA systems in the domain of Islamic literature and provide researchers and practitioners with a valuable resource to explore and understand the Quran, Tafsir, and Ahadith more efficiently and accurately.

\section*{Author Contributions}

\section*{Acknowledgements}
The authors would like to thank the CPInS Research Lab at SEECS-NUST and Prince Sultan University for facilitating the research and publication of this work. The authors also recognize their invaluable support and resources for these efforts.

\bibliographystyle{ACM-Reference-Format}
\bibliography{references}

\section*{Author Biography}
\textbf{FAIZA QAMAR} is a Ph.D scholar at the National University of Sciences and Technology (NUST), Pakistan. She completed her Bachelor’s and master’s in computer science from the University of Gujrat (UOG) and University of Engineering and Technology (UET), respectively. Her research interests are Natural Language Processing (NLP), Machine learning, Deep learning, and Interdisciplinary research.\\

\textbf{SEEMAB LATIF} is an associate professor and a researcher at the National University of Sciences and Technology (NUST), Pakistan. She received her PhD from the University of Manchester, UK. Her research interests include artificial intelligence, machine learning, data mining, and NLP. Her professional services include Industry Consultations, Conference Chair, Technical Program Committee Member, and reviewer for several international journals and conferences. In the last 3 years, she has established research collaborations with national and international universities and institutes. She has also secured grants from the Asian Development Bank, HEDP, World Bank, Higher Education Commission Pakistan, National ICT, HEC Technology Development Fund, and UK ILM Ideas. She received the School Best Teacher award in 2016 and the University Best Innovator Award in 2020. She is also the founder of NUST spin-off company, Aawaz AI Tech.\\

\textbf{RABIA LATIF} is currently working as an Associate Professor at the College of Computer and Information Sciences, Prince Sultan University, Riyadh, Saudi Arabia. She received her bachelor’s degree in Computer Science from COMSATS Institute of Information Technology, Islamabad, and her master’s degree in information security from the National University of Sciences and Technology (NUST), Pakistan. She has done her PhD in cloud-assisted wireless body area networks at the
National University of Sciences and Technology, Pakistan. Her research interests include wireless body area networks, cloud computing, information security and Artificial Intelligence.

\section{Appendices}
\appendix

\section{Answers generated by Finetuned Language Models}
\label{app:ans}
The table presents the answers generated by the language models i.e. T5, BART, LED and LongT5.

 \begin{figure}[h]
     \centering
     \includegraphics[width=\textwidth]{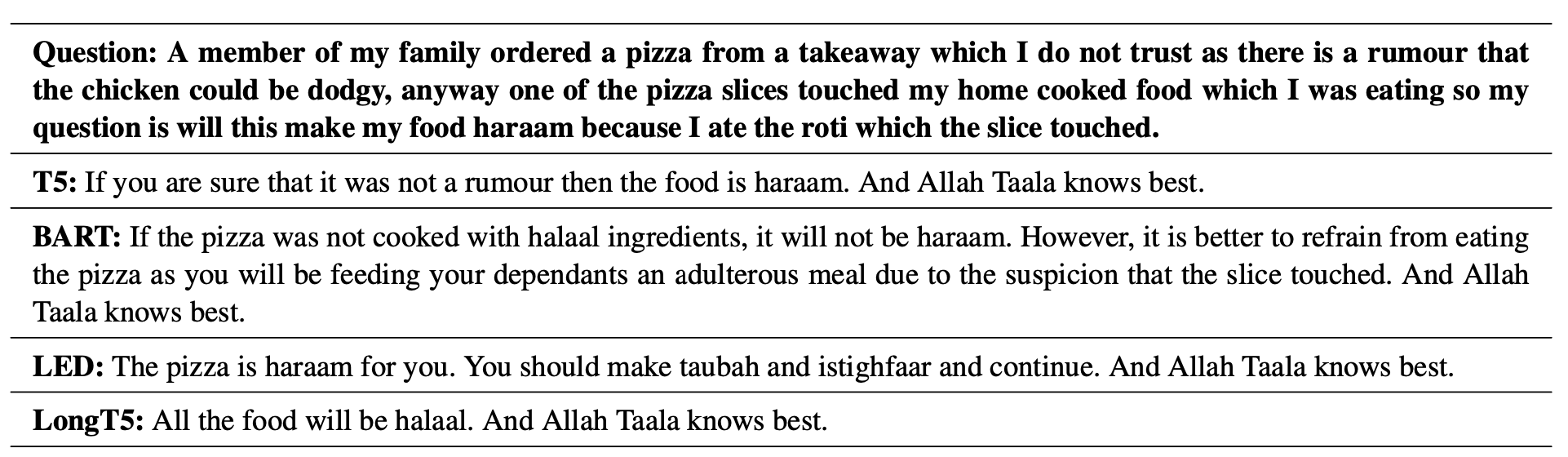}
 \end{figure}

  \begin{figure}[h]
     \centering
     \includegraphics[width=\textwidth]{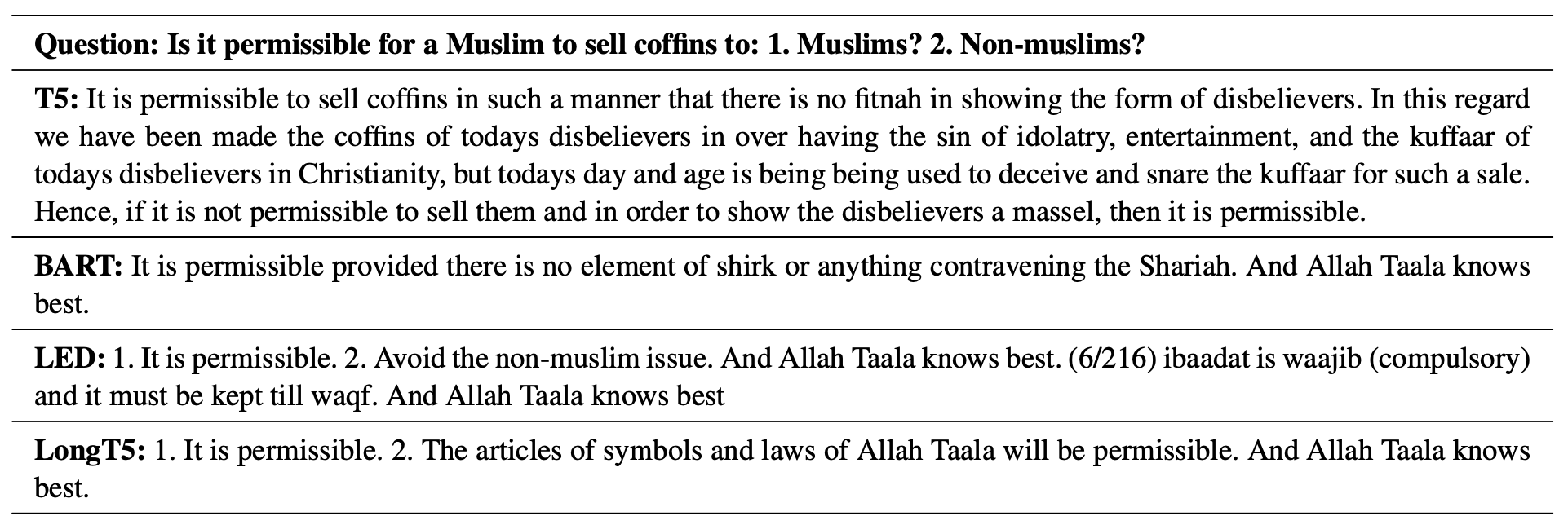}
 \end{figure}

  \begin{figure}[h]
     \centering
     \includegraphics[width=\textwidth]{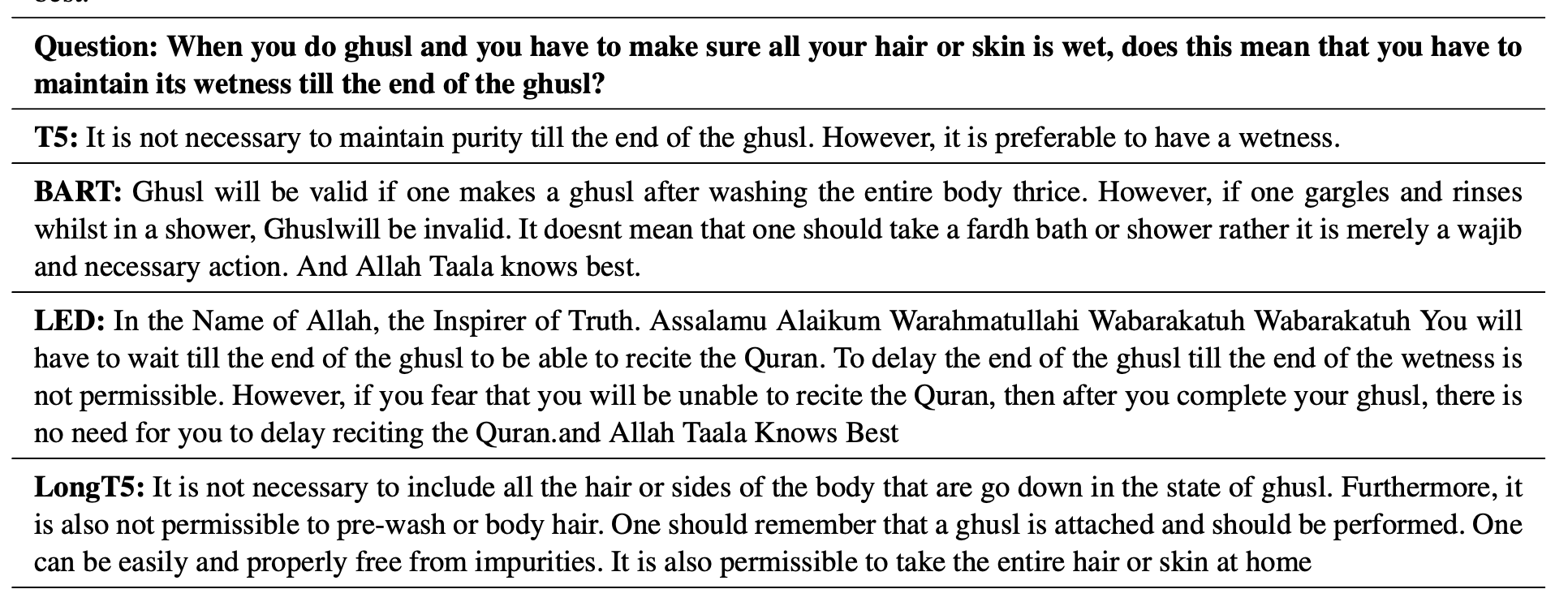}
 \end{figure}

\section{Examples from the Dataset with Context}
\label{app:dataset}
Following are some examples from the dataset. In each example, there are three entities. a) Question, b) Context extracted from Quranic Tafsir and Ahadith, and c) Answer provided by the Muslim Scholars. \\

 \begin{figure}[h]
     \centering
     \includegraphics[width=\textwidth]{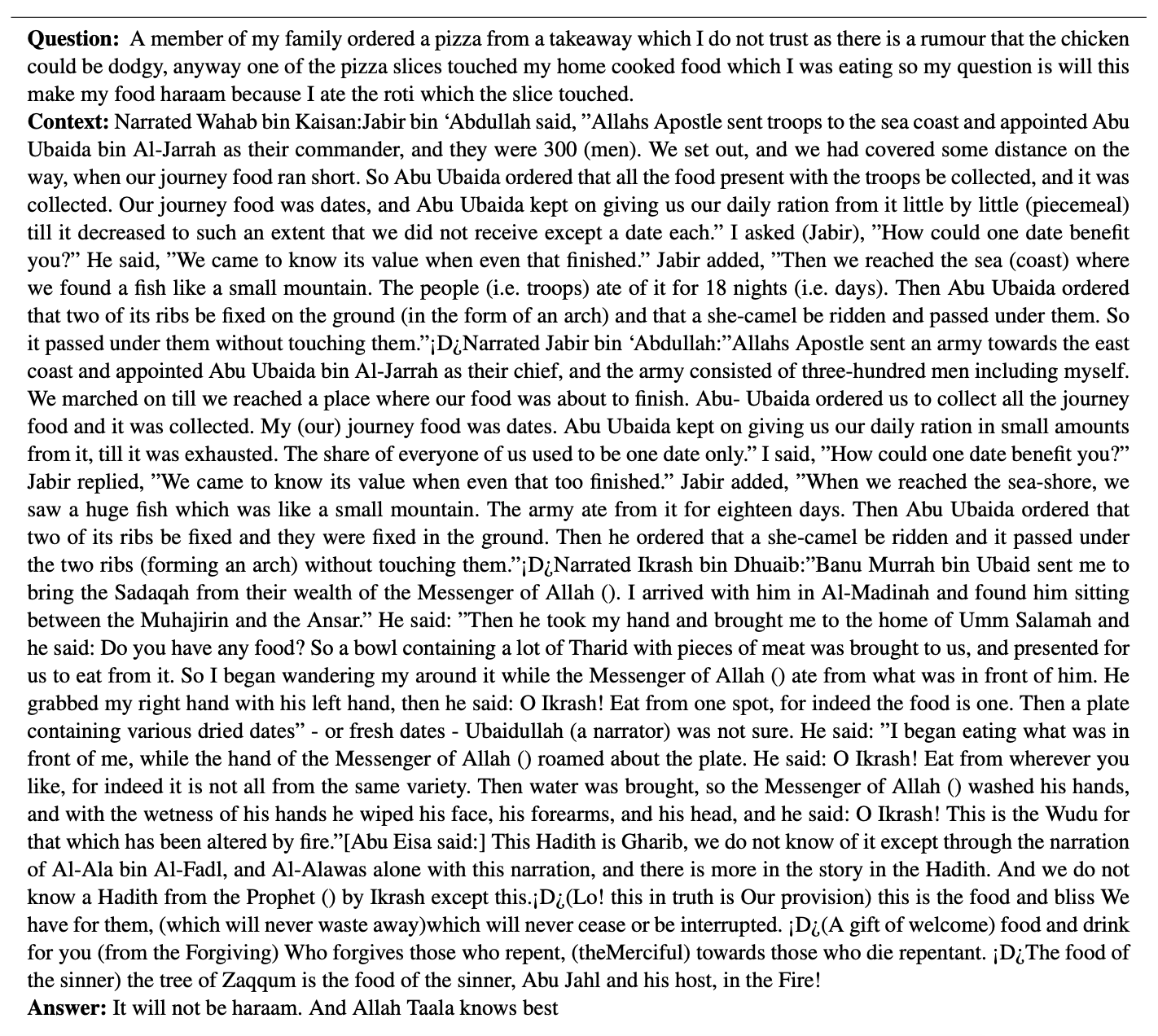}
 \end{figure}

  \begin{figure}[h]
     \centering
     \includegraphics[width=\textwidth]{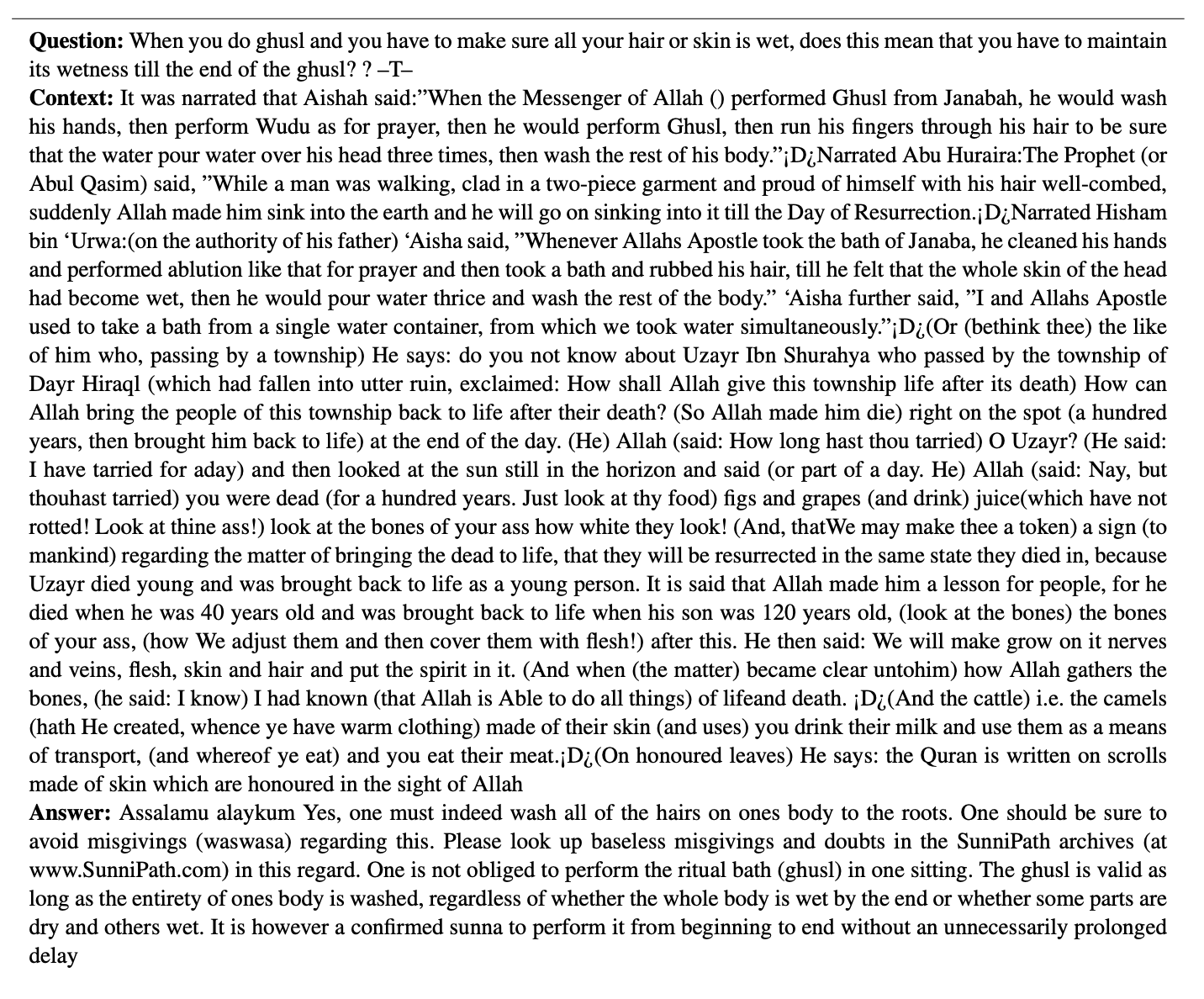}
 \end{figure}

  \begin{figure}[h]
     \centering
     \includegraphics[width=\textwidth]{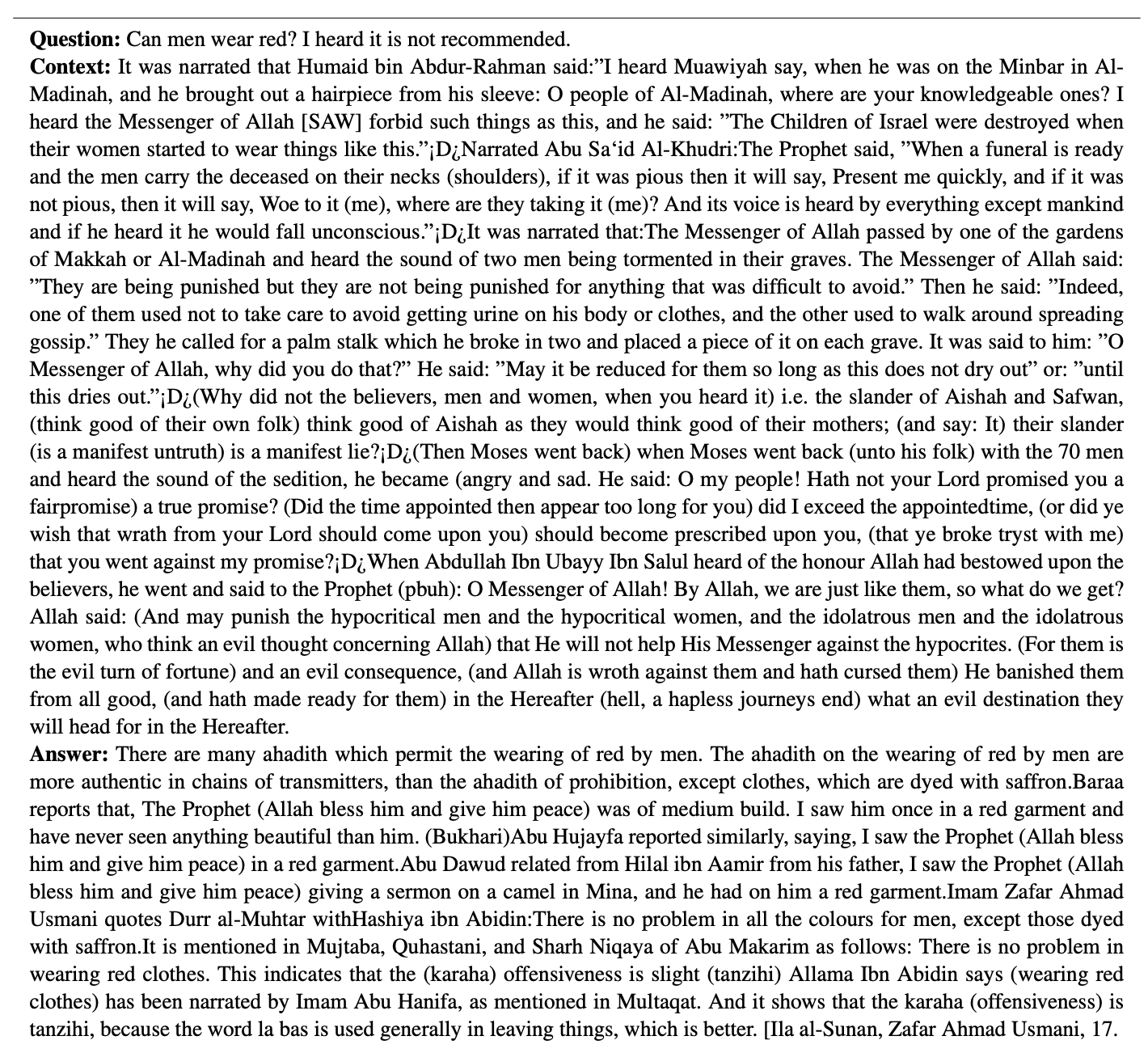}
 \end{figure}
\label{lastpage}

\end{document}